\documentclass[10pt,logo,twocolumn,copyright]{nvidia}

\usepackage[utf8]{inputenc}
\usepackage[T1]{fontenc}

\usepackage{color,xcolor}
\usepackage{epsfig}
\usepackage{graphicx}
\usepackage{duckuments}

\usepackage{stfloats}

\usepackage{adjustbox}
\usepackage{array}
\usepackage{booktabs}
\usepackage{colortbl}
\usepackage{float,wrapfig}
\usepackage{hhline}
\usepackage{multirow}
\usepackage{subcaption}
\usepackage[font=small]{caption}
\usepackage{makecell}
\usepackage{listings}

\usepackage{amsmath,amsfonts,amsthm,amssymb}
\usepackage{bm}
\usepackage{nicefrac}
\usepackage{microtype}

\usepackage{changepage}
\usepackage{extramarks}
\usepackage{fancyhdr}
\usepackage{lastpage}
\usepackage{setspace}
\usepackage{soul}
\usepackage{xspace}
\usepackage{indentfirst}
\usepackage{pifont}
\usepackage{cuted}
\usepackage{wrapfig}
\definecolor{cvprblue}{rgb}{0.21,0.49,0.74}

\usepackage[pagebackref,breaklinks,colorlinks,citecolor=cvprblue]{hyperref}
\usepackage{url}

\usepackage{algorithm, algorithmic}
\usepackage{enumitem}

\usepackage{wasysym}
\usepackage{todonotes}
\usepackage{pifont}
\usepackage{fancyvrb}
\usepackage{fvextra}
\newcolumntype{L}[1]{>{\raggedright\let\newline\\\arraybackslash\hspace{0pt}}m{#1}}
\newcolumntype{C}[1]{>{\centering\let\newline\\\arraybackslash\hspace{0pt}}m{#1}}
\newcolumntype{R}[1]{>{\raggedleft\let\newline\\\arraybackslash\hspace{0pt}}m{#1}}

\newcommand{\sect}[1]{\S~\ref{sect:#1}}

\newcommand{\fig}[1]{Figure~\ref{fig:#1}}

\newcommand{\tbl}[1]{Table~\ref{tab:#1}}

\newcommand{\lblfig}[1]{\label{fig:#1}}
\newcommand{\lblsect}[1]{\label{sect:#1}}

\newcommand{\lbltbl}[1]{\label{tab:#1}}

\newcommand{\ignorethis}[1]{}

\makeatletter
\DeclareRobustCommand\onedot{\futurelet\@let@token\@onedot}
\def\@onedot{\ifx\@let@token.\else.\null\fi\xspace}

\makeatother

\definecolor{citecolor}{rgb}{34,139,34}
\definecolor{mydarkblue}{rgb}{0,0.08,1}
\definecolor{mydarkgreen}{rgb}{0.12,0.7,0.12}
\definecolor{mydarkred}{rgb}{0.8,0.02,0.02}
\definecolor{mydarkorange}{rgb}{0.40,0.2,0.02}
\definecolor{mypurple}{RGB}{111,0,255}
\definecolor{myred}{rgb}{1.0,0.0,0.0}
\definecolor{mygold}{rgb}{0.75,0.6,0.12}
\definecolor{mydarkgray}{rgb}{0.66,0.66,0.66}

\definecolor{darkgreen}{rgb}{0.15, 0.75, 0.15}
\definecolor{mitblue}{rgb}{0.88,0.95,0.96}
\definecolor{lightblue}{rgb}{0.90, 0.95, 0.99}

\newcommand{\method}{Jet-RL}
\newcommand{\naivefpeight}{BF16-train-FP8-rollout}
\newcommand{\NaiveFpEight}{BF16-Train-FP8-Rollout}

\newcommand{\LongRollout}{long-rollout generation}
\newcommand{\ChallengingTask}{challenging tasks}

\newcommand{\gain}[1]{\textcolor{green!60!black}{\scriptsize\, +#1$\uparrow$}}
\newcommand{\loss}[1]{\textcolor{red!70!black}{\scriptsize\, -#1$\downarrow$}}

\usepackage{amsmath,amsfonts,bm}

\def\eqref#1{equation~\ref{#1}}

\def\1{\bm{1}}

\def\mW{{\bm{W}}}
\def\mX{{\bm{X}}}
\def\mY{{\bm{Y}}}

\DeclareMathAlphabet{\mathsfit}{\encodingdefault}{\sfdefault}{m}{sl}
\SetMathAlphabet{\mathsfit}{bold}{\encodingdefault}{\sfdefault}{bx}{n}

\def\gE{{\mathcal{E}}}

\def\gG{{\mathcal{G}}}

\def\gV{{\mathcal{V}}}

\def\sR{{\mathbb{R}}}

\ifx\theorem\undefined
\newtheorem{theorem}{Theorem}[section]
\fi

\ifx\example\undefined
\newtheorem{example}{Example}[section]
\fi

\ifx\property\undefined
\newtheorem{property}{Property}
\fi

\ifx\lemma\undefined
\newtheorem{lemma}{Lemma}[section]
\fi

\ifx\proposition\undefined
\newtheorem{proposition}{Proposition}[section]
\fi

\ifx\remark\undefined
\newtheorem{remark}{Remark}
\fi

\ifx\corollary\undefined
\newtheorem{corollary}{Corollary}[section]
\fi

\ifx\definition\undefined
\newtheorem{definition}{Definition}[section]
\fi

\ifx\conjecture\undefined
\newtheorem{conjecture}{Conjecture}[section]
\fi

\ifx\axiom\undefined
\newtheorem{axiom}[theorem]{Axiom}
\fi

\ifx\claim\undefined
\newtheorem{claim}[theorem]{Claim}
\fi

\ifx\assumption\undefined
\newtheorem{assumption}{Assumption}
\fi

\ifx\problem\undefined
\newtheorem{problem}{Problem}[section]
\fi

\ifx\fact\undefined
\newtheorem{fact}{Fact}
\fi

\begin{document}

\title{Jet-RL: Enabling On-Policy FP8 Reinforcement Learning with Unified Training and Rollout Precision Flow}

\author{
Haocheng Xi\textsuperscript{1,3} 
Charlie Ruan\textsuperscript{3} 
Peiyuan Liao\textsuperscript{4} 
Yujun Lin\textsuperscript{1} 
Han Cai\textsuperscript{1} 
Yilong Zhao\textsuperscript{3} 
Shuo Yang\textsuperscript{3} 
Kurt Keutzer\textsuperscript{3} 
Song Han\textsuperscript{1,2} 
Ligeng Zhu\textsuperscript{1,$\dag$} \qquad 
~\\
\textsuperscript{1}NVIDIA \quad \textsuperscript{2}MIT \quad \textsuperscript{3}UC Berkeley \quad \textsuperscript{4} Independent Researcher 
}

\begin{abstract}
Reinforcement learning (RL) is essential for enhancing the complex reasoning capabilities of large language models (LLMs). However, RL training pipelines are computationally inefficient and resource-intensive, with the rollout phase consuming over 70\% of the total training time. 
Quantized RL training, particularly with FP8, offers a promising solution. For example, a common approach is to employ FP8 precision during rollout to alleviate this bottleneck while retaining BF16 precision during training. 
In this work, we present the first comprehensive study of FP8 RL training and show that the commonly adopted \textit{BF16-train + FP8-rollout} strategy suffers from severe training instability and catastrophic accuracy collapse under \LongRollout{} and \ChallengingTask{}. Our analysis reveals that these issues arise from the off-policy nature of the approach, which introduces significant numerical mismatch between training and inference.
Motivated by these findings, we propose \method{}, a FP8 RL training framework that enables robust and stable RL training. The key idea is to adopt \textbf{an unified FP8 precision flow for both training and rollout}, minimizing numerical discrepancies and avoiding the need for inefficient inter-step calibration. 
Extensive experiments demonstrate the effectiveness of \method{}. Our method achieves up to 33\% rollout phase speedup, up to 41\% training phase speedup, and a 16\% end-to-end speedup over BF16 training while maintaining robust convergence across all settings and exhibiting negligible accuracy degradation. We will release our code and pre-trained models when less anoynomous.
\end{abstract}

%

\maketitle

\section{Introduction}

\begin{figure}[t]
    \centering
    \includegraphics[width=\linewidth]{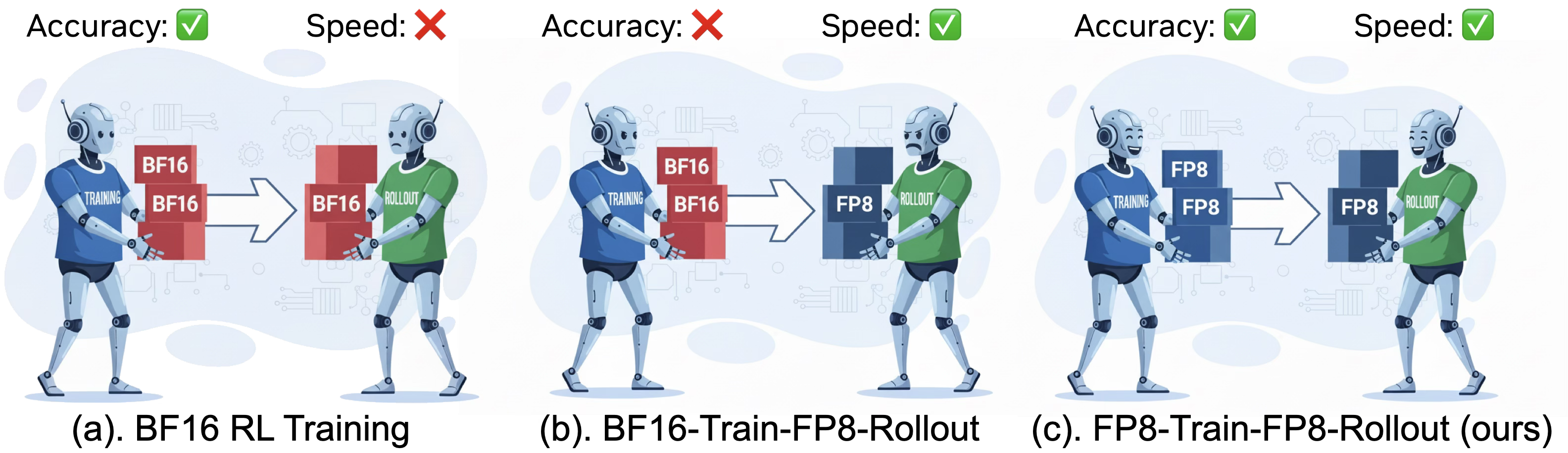}
    \caption{\textbf{Overview of RL training different between JetRL and other methods.} JetRL proposes unified precision flow for FP8 RL training accommodate both performance and throughput .}
    \lblfig{teaser}
\end{figure}
\begin{figure}[t]
    \centering
    \includegraphics[width=\linewidth]{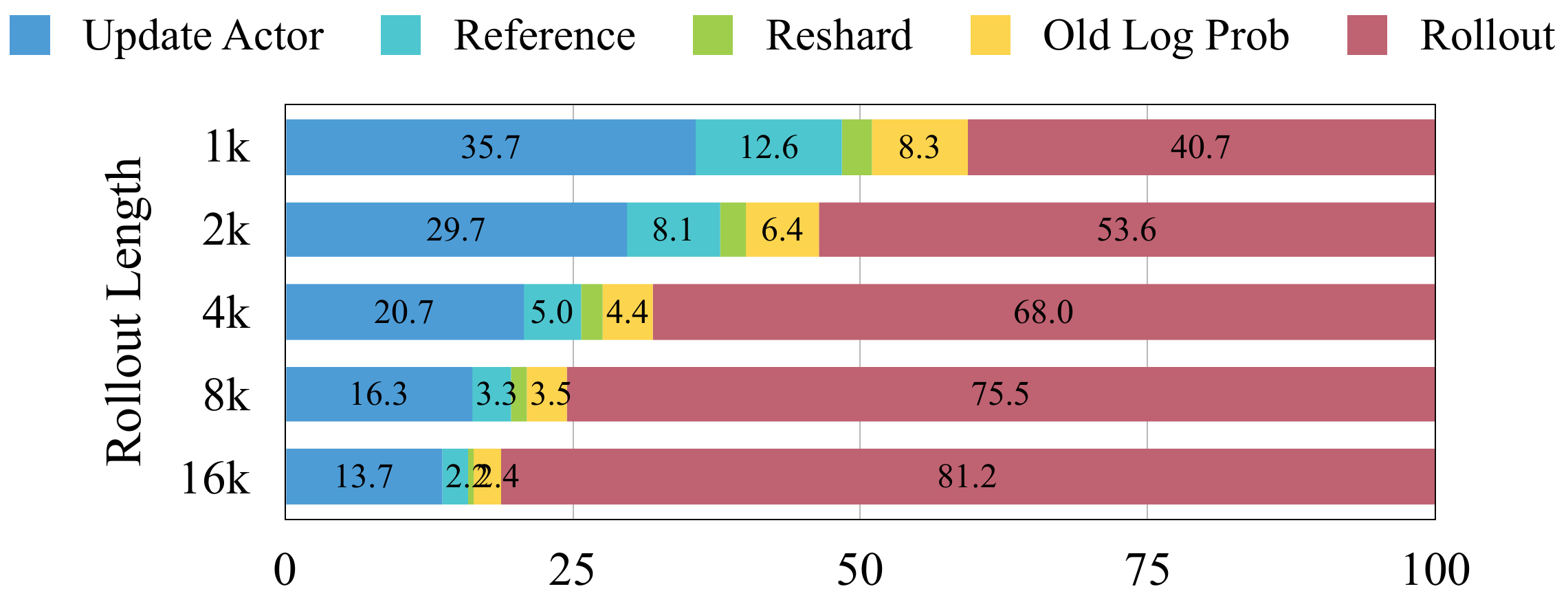}
    \caption{\textbf{Rollout Generation dominates the RL training latency.} When the rollout length is larger than 8k, rollout will take $>75\%$ of the total latency, making it the primary bottleneck.}
    \lblfig{rollout_is_slow}
\end{figure}
\begin{figure*}[t]
    \centering
    \includegraphics[width=0.9\linewidth]{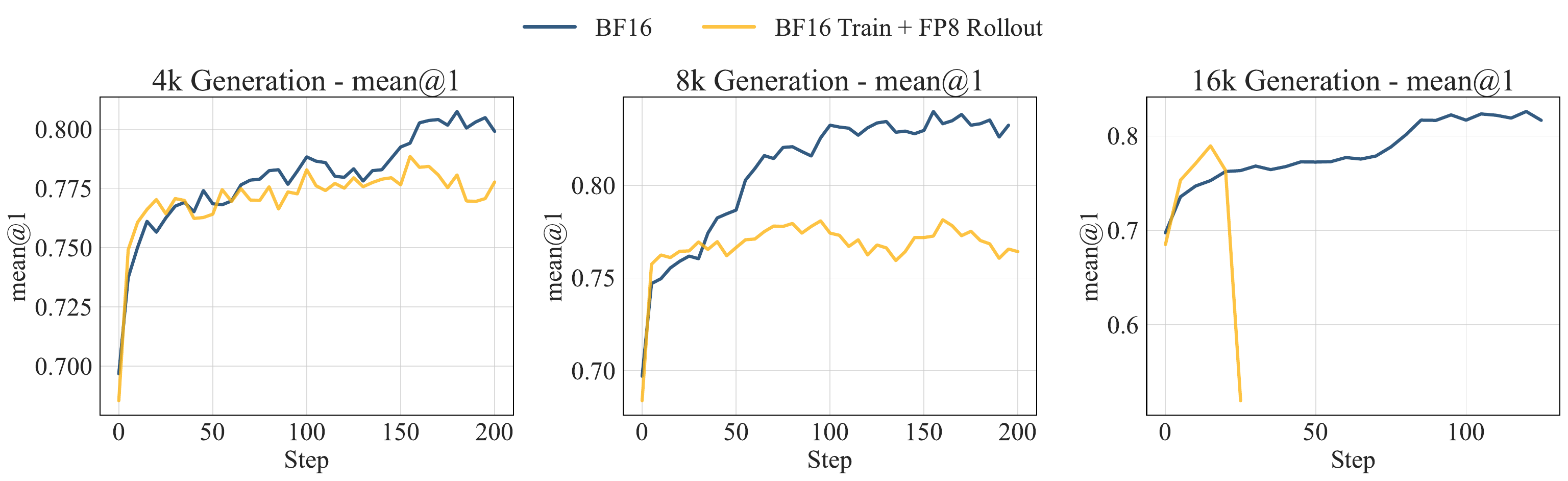}
    \caption{\textbf{Naive BF16 Train + FP8 Rollout fails when rollout context length increases. }We observe that although the \NaiveFpEight{}  method might exhibit similar performance compared with BF16 training, its performance quickly degrades as we extend the rollout length to more than 8k. We use the Qwen3-8B-Base model for this experiment, and evaluate and train on MATH.}
    \lblfig{long_rollout_diverge}
\end{figure*}

The emergence of frontier reasoning large language models (LLMs)~\cite{jaech2024openai,guo2025deepseek} has demonstrated unprecedented capabilities in solving highly complex, multi-step problems~\cite{chollet2024arc,chollet2025arcagi2newchallengefrontier}, setting new state-of-the-art across domains ranging from advanced mathematics to scientific discovery. A key driver of this success is the Chain-of-Thought (CoT)~\cite{zhou2024chainofthoughtreasoningwithout,Wei2022Chainofthought} paradigm, which enables models to generate extended, coherent reasoning traces before producing final answers. This mechanism allows LLMs to conduct detailed analyses, explore alternative solution paths, and execute structured logical reasoning. Reinforcement Learning (RL)~\cite{Ouyang2022TrainingRLHF,Bai2022TrainingRLHF2} has emerged as the pivotal training paradigm for unlocking robust CoT generation, propelling models from simple question answering toward goal-directed reasoning.

However, RL training remains notoriously resource-intensive, with the actor rollout phase becoming the primary training bottleneck~\cite{yu2025dapoopensourcellmreinforcement}. Effective reasoning often demands generating long token sequences (e.g., $>$6,000 tokens) to adequately explore the solution space. Owing to the inherently autoregressive nature of LLMs, this rollout stage incurs substantial training time, frequently accounting for more than 70\% of the total end-to-end training pipeline (\fig{rollout_is_slow}). This extreme imbalance significantly hampers overall training efficiency, making rollout acceleration an urgent priority~\cite{Jin2025SearchR1TL,cheng2025Reasoning360}.

To improve efficiency, \textbf{FP8 quantization} offers a highly promising solution. It has been widely adopted in LLM inference, demonstrating significant efficiency gains~\cite{fishman2025scalingfp8trainingtrilliontoken,peng2023fp8lm,xi2024coat,kim2025inquiryfp8}. A natural approach to accelerate RL training is to apply FP8 quantization during the rollout phase, which constitutes the main efficiency bottleneck, and retain BF16 precision for the training phase, which is expected to deliver better training stability and accuracy. This strategy\textit{~\naivefpeight{}} has been widely adopted in modern RL frameworks such as VeRL, SLIME, Nemo-RL, and OpenRLHF~\cite{sheng2024hybridflow_verl, slime_github, nemo-rl, hu2024openrlhf}. However, we find that this strategy exhibits critical limitations in two circumstances: 


\paragraph{(i) Long-Rollout Generation.} First, we observe that this training strategy suffers from significant accuracy degradation and can even experience catastrophic collapse when applied to long sequences. For example, Figure~\ref{fig:long_rollout_diverge} shows the training curves of the Qwen3-8B-Base model across 4K, 8K, and 16K token generations. Compared with BF16 training, its accuracy rapidly drops once the rollout length exceeds 8K tokens, ultimately resulting in significant performance drop at the 16K setting.

Small numerical discrepancies will gradually accumulate during long chain-of-thought reasoning. The mismatch between training and rollout precision are negligible for short sequences but progressively increases with the rollout length. These accumulated errors amplify the effects of off-policy training, causing the generation trajectory to diverge and making RL training unstable. 

\paragraph{(ii) Challenging Tasks.} Second, we observe that the claimed success of recent \naivefpeight{} is not always reliable. As shown in Figure~\ref{fig:WeakerModelFail}, the strategy shows no degradation when the model already possesses strong task priors from pretraining. However, when applied to \textit{harder reasoning tasks} or trained with \textit{weaker base models}, the training curve quickly diverges between BF16 and FP8 rollouts. These observations suggest that effective quantized rollout must be robust and adaptable across diverse training settings.

To address these limitations and establish FP8 rollout as a reliable acceleration strategy, we propose \method{}. Our core contribution is enforcing a \textbf{truly on-policy FP8 training paradigm} that stabilizes RL training. We design our framework to use an \textbf{identical quantization precision flow} for both training and inference, eliminating policy mismatch and removing the need for inter-step calibration. Our approach adopts a mixed per-group and per-block quantization scheme~\cite{cheng2025deepseekv3technicalreport,xi2024coat} and leverages state-of-the-art FP8 GEMM kernels~\cite{deepseek2025deepgemm} to achieve acceleration for end-to-end RL training. 

We conduct comprehensive experiments across diverse models, datasets, and rollout configurations to validate \method{}. Our results demonstrate that \method{} successfully stabilizes training, minimizes divergence between training and rollouts, and substantially narrows the performance gap between BF16 and FP8 RL. While \naivefpeight{} methods typically incur more than $5\%$ performance degradation compared to BF16 baselines, our approach reduces this to $\sim 1\%$. Meanwhile, \method{} achieves up to $1.33\times$ rollout phase speedup for 32B model, up to $1.41\times$ training phase speedup for 8B model, and $1.16\times$ end-to-end speedup for 8B experiments. These findings confirm that our method provides a robust solution for efficient low-precision RL training, enabling significant acceleration without sacrificing performance.

We summarize our contributions below: 
\begin{itemize}[leftmargin=24pt,topsep=0pt]
    \item We identify that the commonly used \naivefpeight{} paradigm leads to training instability and accuracy collapse under \LongRollout{} and \ChallengingTask{}. 
    \item We propose \method{}, an on-policy FP8 RL framework by enforcing an unified training–rollout precision flow for training and rollout. This approach resolves policy mismatch and ensures stable optimization.
    \item \method{} achieves substantial rollout and end-to-end speedups while maintaining convergence and accuracy close to BF16 baselines across various models and tasks.
\end{itemize}

\section{Background}

\subsection{Quantization Basis}
Quantization maps a high-precision tensor to a lower-precision tensor to speed up computation and reduce memory footprint. We consider quantizing a tensor $\mX$ into a target data format whose maximum representable value is $\Delta_{\text{max}}$. The quantization process can be defined as:
\[
    \hat{\mX}, S_{\mX} = Q(\mX), \text{ where } Q(\cdot) \text{ is the quantizer}
\]
\[
    \hat{\mX} = \left\lceil \frac{\mX}{S_\mX} \right\rfloor, \quad
    S_\mX = \frac{\max(|\mX|)}{\Delta_{\text{max}}}.
\]

Here, $\hat{\mX}$ is the low-precision tensor obtained after quantization, and $S_\mX$ is the scaling factor. In this paper, we focus on FP8 quantization using the E4M3 format~\cite{micikevicius2022fp8,micikevicius2023ocp}, whose maximum representable value is $\Delta_{\text{max}}=448$.

\subsection{Quantization of a Linear Layer}\lblsect{quantization_linear_layer}
When quantizing large language models, linear layers are the primary operators of interest, as they are compute-bound. Following previous works on quantized training~\cite{xi2024jetfire}, the input and output of a linear layer are denoted as:
\[
\mY \in \sR^{N \times D}, \quad \mX \in \sR^{N \times C}, \quad \mW \in \sR^{D \times C},
\]
where $\mY$ is the output, $\mX$ is the activation, and $\mW$ is the weight. $N$ is the number of tokens, $D$ is the number of output channels, and $C$ is the number of input channels. Their corresponding gradients are denoted as $\nabla_\mY, \nabla_\mX, \text{and } \nabla_\mW$, each with the same shape. 

Each linear layer has three GEMMs: \textit{FProp} in the forward pass, \textit{WGrad} in the backward pass to calculate the gradient of weights, and \textit{DGrad} in the backward pass to calculate the gradient of activations. Formally, they can be expressed as:

\textbf{Forward pass (FProp)}
\[
\mY = \mX \times \mW^{\top}, \quad \text{where} \quad \sR^{N \times D} = \sR^{N \times C} \times \sR^{C \times D}.
\]

\textbf{Backward pass -- Gradient of weights (WGrad)}
\[
\nabla_{\mW} = \nabla_{\mY}^{\top} \times \mX, \quad \text{where} \quad \sR^{D \times C} = \sR^{D \times N} \times \sR^{N \times C}.
\]

\textbf{Backward pass -- Gradient of input (DGrad)}
\[
\nabla_{\mX} = \nabla_{\mY} \times \mW, \quad \text{where} \quad \sR^{N \times C} = \sR^{N \times D} \times \sR^{D \times C}.
\]

Quantizing these three GEMMs for FP8 computation has specific matrix layout requirements, as current FP8 TensorCore hardware expects the first operand to be stored in a row-wise (R) manner and the second operand in a column-wise (C) manner. The layout  are summarized in \tbl{fp8_layouts}.

\begin{table}[!h]
\centering
\captionof{table}{Layout requirements for FP8 GEMMs in linear layer.}
\lbltbl{fp8_layouts}
\vspace{5pt}
\begin{tabular}{c|ccccc}
\toprule
GEMM & $\mX$ & $\mW$ & $\nabla_{\mY}$ & Expression\\
\midrule
\textbf{FProp} & Row & Row & -- & $\mX \times \mW^{\top}$ \\
\textbf{WGrad} & Col & - & Col & $\nabla_{\mY}^{\top} \times \mX$ \\
\textbf{DGrad} & - & Col & Row & $\nabla_{\mY} \times \mW$ \\
\bottomrule
\end{tabular}
\end{table}

\subsection{Workload of Reinforcement Learning}

A standard Reinforcement Learning (RL) training pipeline, such as one based on Proximal Policy Optimization (PPO)~\cite{schulman2017proximalpolicyoptimizationalgorithms}, typically consists of four distinct models. The \textit{Actor Model} is the primary Large Language Model (LLM) we aim to train. The \textit{Reference Model}, often an initial copy of the actor, is used to calculate the Kullback–Leibler (KL) divergence, which regularizes the actor's updates and ensures training stability. The \textit{Reward Model} provides the scalar reward signal, typically learned to reflect human preferences or task objectives. Finally, the \textit{Critic Model} estimates the value (or quality) of the generated responses, predicting expected future rewards.

Each RL training step can be broken into three distinct phases, where each phase is a composite of three fundamental LLM workloads: \textit{decode}, \textit{prefill}, and \textit{training}.
First, in the \textbf{Rollout} phase, the Actor model performs an autoregressive \textit{decode} process to generate one or more responses for a given prompt. We use the term ``rollout'' for this generation stage to differentiate it from simple inference.
Second, in the \textbf{Evaluation} phase, the generated responses are fed into the other models. The Reference, Reward, and Critic models each perform a single forward pass, which constitutes a \textit{prefill} workload, to compute their respective outputs.
Third, in the \textbf{Update} phase, these metrics are gathered (often processed via Generalized Advantage Estimation (GAE)~\cite{schulman2018highdimensionalcontinuouscontrolusing}), and the Actor model executes a \textit{training} step, consisting of one forward and one backward pass, to update its weights.

These phases have distinct computational profiles, leading to a common implementation strategy that employs two different types of systems. The Rollout phase is typically managed by optimized inference engines like vLLM~\cite{Kwon2023EfficientMM} or SGLang~\cite{Zheng2023SGLangEE}, as they are highly optimized for autoregressive workloads. Conversely, the Evaluation and Update phases are handled by training frameworks such as FSDP~\cite{Zhao2023PyTorchFE}, Megatron-LM~\cite{Shoeybi2019MegatronLMTM}, or DeepSpeed~\cite{rasley2020deepspeed}, which provide greater flexibility in parallelism. To maintain the RL training on-policy , the weights of updated actor must be transferred from the training framework to the inference engine after each step.

Despite optimizations from inference engines, the rollout stage remains a critical performance bottleneck. As we show in \sect{rollout_is_slow}, the rollout phase latency scales with the length of the generated response. Consequently, this stage gradually dominates the end-to-end training latency, making it the most expensive part of the entire RL training pipeline.

\section{Motivation}

\subsection{Rollout is the Bottleneck in RL Training}\lblsect{rollout_is_slow}

In \fig{rollout_is_slow}, we profile each component in RL training and experiment with the Qwen3-8B-Base~\cite{yang2025qwen3} model on GSM8K~\cite{Cobbe2021GSM8k} and MATH~\cite{hendrycks2021measuringmathematicalproblemsolving}. To understand how training time scales with rollout length, we vary the maximum rollout length from 1K to 16K and measure how the latency proportion of each training component changes. We observe that when the rollout length exceeds 8K, rollout generation alone accounts for over 70\% of the total training time. This trend is consistent with observations reported in previous studies~\cite{yu2025dapoopensourcellmreinforcement}, highlighting the critical need to accelerate rollout. 

FP8 quantization is a promising solution to speed up rollout for two reasons. First, it can offer an ideal $2\times$ speedup over BF16. Second, a wide range of studies shows that FP8 inference does not degrade performance on downstream tasks. Moreover, it can be easily integrated into existing RL training pipelines. 
Existing studies have proposed simply casting the BF16 weights to FP8 while keeping BF16 for the training phase. In the rest of this paper, we refer to this simple strategy as \textit{\naivefpeight{}} and discuss the limitations of such design.


\subsection{\NaiveFpEight{} with Calibration is Slow }

While most Post-Training Quantization (PTQ) methods~\cite{Lin2023AWQ,Frantar2022GPTQ, Xiao2022SmoothQuantAA} are designed for offline deployment, RL training requires \textbf{frequent weight synchronization} between the training actors and the rollout actors. 
The expensive and data-dependent calibration can take tens of minutes even for small 8B LLMs, thus becoming unaffordable and conflicting with the purpose of acceleration if repeated at every synchronization step. Some frameworks such as 
SLIME~\cite{SLIME_FP8Rollout} and NeMo-RL~\cite{Nemo-RL_FP8Rollout} propose to directly cast the BF16 weights to FP8 without calibration and claim the accuracy is not affected. However, we find this conclusion is in fact very fragile.


\subsection{\NaiveFpEight{} without Calibration is Unstable}

The quality of RL training highly relies on the \textbf{on-policy assumption}, which dictates that the agent learns from data and experiences collected while following its current policy---in LLM training, this requires that rollout and training should produce the same logits when given the same prompts. 
RL training without this consistency may lead to divergence in extreme cases~\cite{he2025nondeterminismThinky,yuan2025GiveMeFP32,sglang_deterministic_2025}. 
Directly quantizing the rollout actor to FP8 while updating a full-precision actor in BF16 clearly breaks this consistency and introduces significant policy mismatch~\cite{yao2025offpolicy,zheng2025prosperitycollapsefaroffpolicy,cetin2022stabilizingoffpolicydeepreinforcement}. 
%

Though in some cases and some studies like Truncated Importance Sampling (TIS), \naivefpeight{} does not result in much performance degradation, we find that this conclusion is very fragile and depends on specific dataset and model settings. Specifically, we observe that \naivefpeight{} tends to fail more in two scenarios: \textit{\LongRollout{}} and \textit{\ChallengingTask{}}. 

\begin{figure}[t]
\centering
\vspace{10pt}
\includegraphics[width=\linewidth]{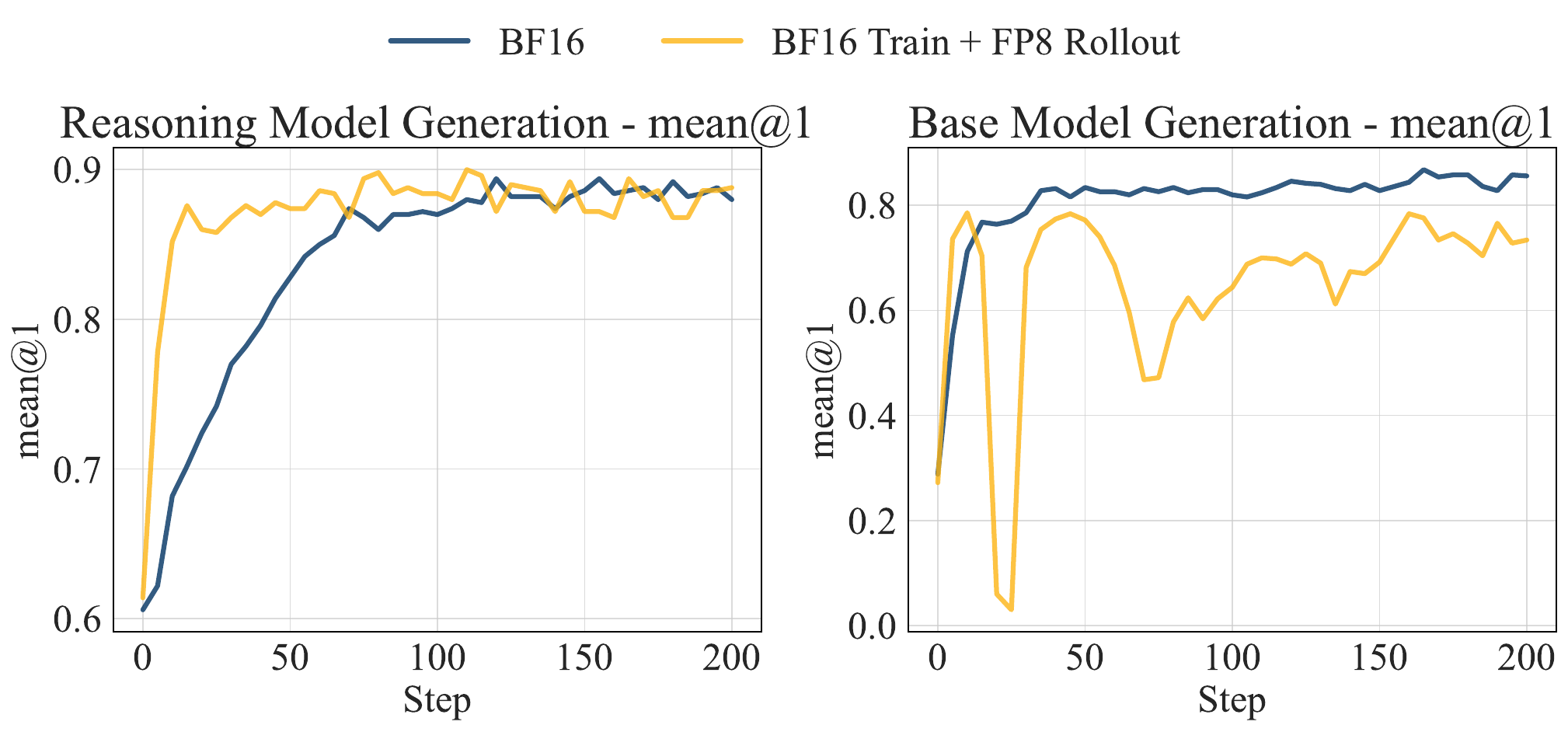}
\caption{We conduct training using Qwen3-8B (reasoning) and Qwen3-8B-Base on MATH. When the model is trained on a task that is easy for itself, \NaiveFpEight{} is less likely to degrade. When the task is hard for the model we trained on, degradation is likely to happen. }
\lblfig{WeakerModelFail}
\end{figure}

\textbf{Failure in \LongRollout{}.}
We observe that the performance of \naivefpeight{} is strongly correlated with rollout length. We train and evaluate Qwen2.5-7B~\cite{qwen2025qwen25technicalreport} on MATH using GRPO, varying the generation length from 4K to 16K. We then compare the performance of BF16 training and \naivefpeight{}. 

As shown in \fig{long_rollout_diverge}, although \naivefpeight{} performs on par with BF16 training when the rollout length is small ($<$4K), the FP8 rollout quickly diverges from BF16 training when rollout length scales to 8K. When we further increase the rollout length to 16K, FP8 rollout's accuracy collapses after only 20 steps of training. 

We hypothesize that this degradation arises from the accumulation of differences between the rollout and training distributions at each decoding step. For shorter generation lengths, these discrepancies remain relatively minor and may even be partially beneficial for training. However, as the rollout length increases, the cumulative divergence between the rollout and training distributions becomes substantial. This intensifies the off-policy issue in reinforcement learning under FP8 rollouts, leading to significant instability and degraded training performance.

\textbf{Failure in \ChallengingTask{}.} 
We observe that \naivefpeight{} tends to fail when the underlying model lacks strong capability on the target task. For instance, as shown in \fig{WeakerModelFail}, when training Qwen3-8B on GSM8K, the \naivefpeight{} training curve closely matches the BF16 training curve and even converges faster. However, after switching to the Qwen3-8B-Base model, which has not undergone instruction tuning and needs to learn both reasoning patterns and question-solving ability, \naivefpeight{} soon falls behind the BF16 baseline and leads to inferior performance. In this setting, the evaluation accuracy diverges from the BF16 baseline after only 20 training steps.

We hypothesize that \naivefpeight{} performs well when the \textit{model is strong} and the \textit{task is relatively simple}. In such cases, the model exhibits high confidence in its responses, rendering it less sensitive to minor numerical perturbations brought by FP8 quantization. Conversely, as task difficulty increases and the model's confidence decreases, quantization-induced errors can substantially distort the rollout trajectory. This leads to a growing mismatch between the FP8 rollout and BF16 training, resulting in unstable optimization and degraded performance. In essence, \naivefpeight{} is only robust for easier tasks but faces convergence issues when scaling to harder tasks. 
Considering that the goal of RL is to help models acquire abilities they don't yet have, \naivefpeight{} is unlikely to suffice, as its instability on harder tasks hinders effective learning and scalability.


These observations motivate a key question: How can we mitigate the training–inference discrepancy in \naivefpeight{} to achieve competitive and stable performance across varying settings?

\section{\method{}: Enabling on-policy FP8 RL Training}

\begin{figure*}[t]
    \centering
    \includegraphics[width=\linewidth]{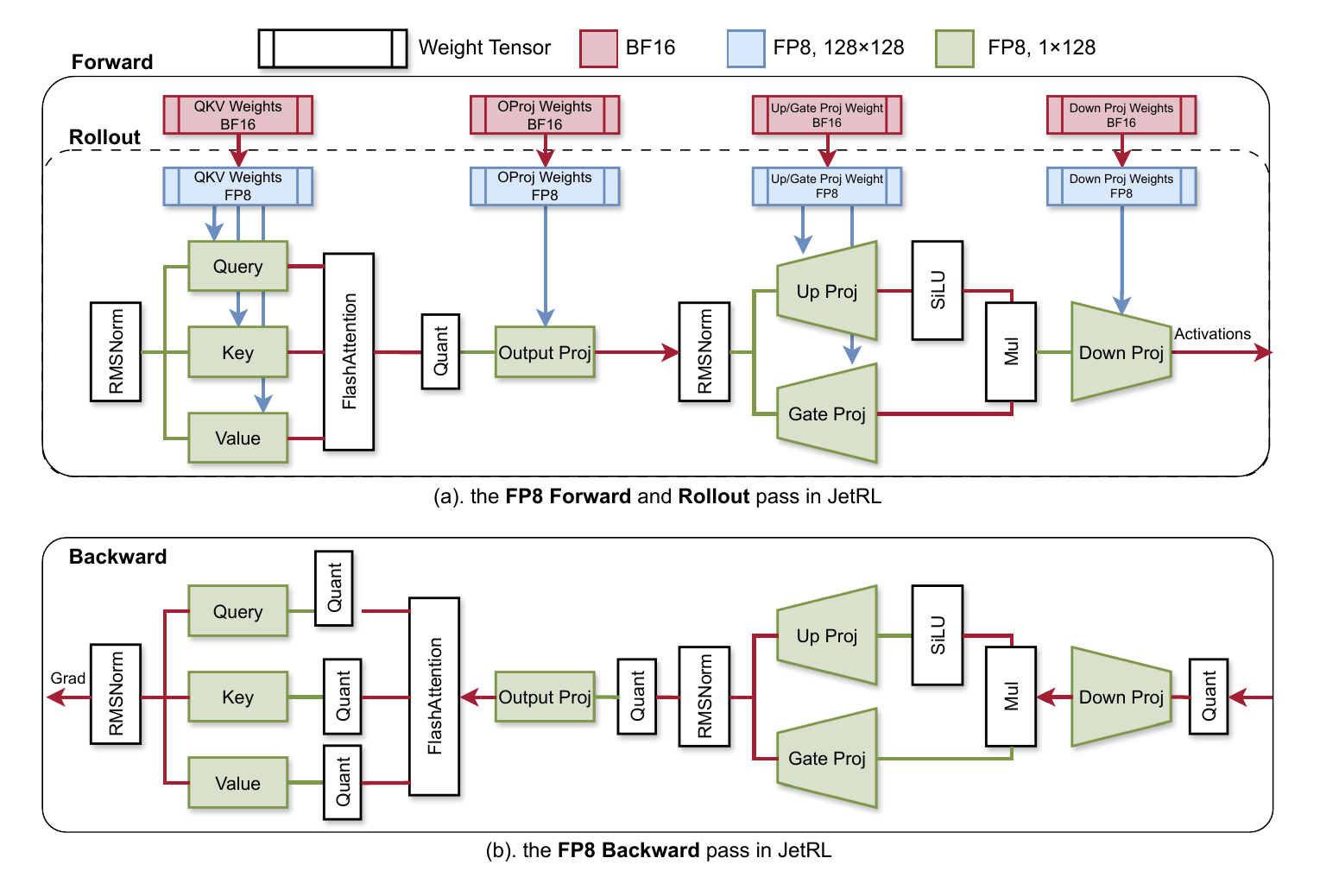}
    \caption{\textbf{Overall FP8 precision flow of \method{}}. 
    The inference graph is a subgraph of the training graph.
    }
    \lblfig{quantization_precision_flow}
\end{figure*}

\begin{figure*}[t]
    \centering
    \includegraphics[width=\linewidth]{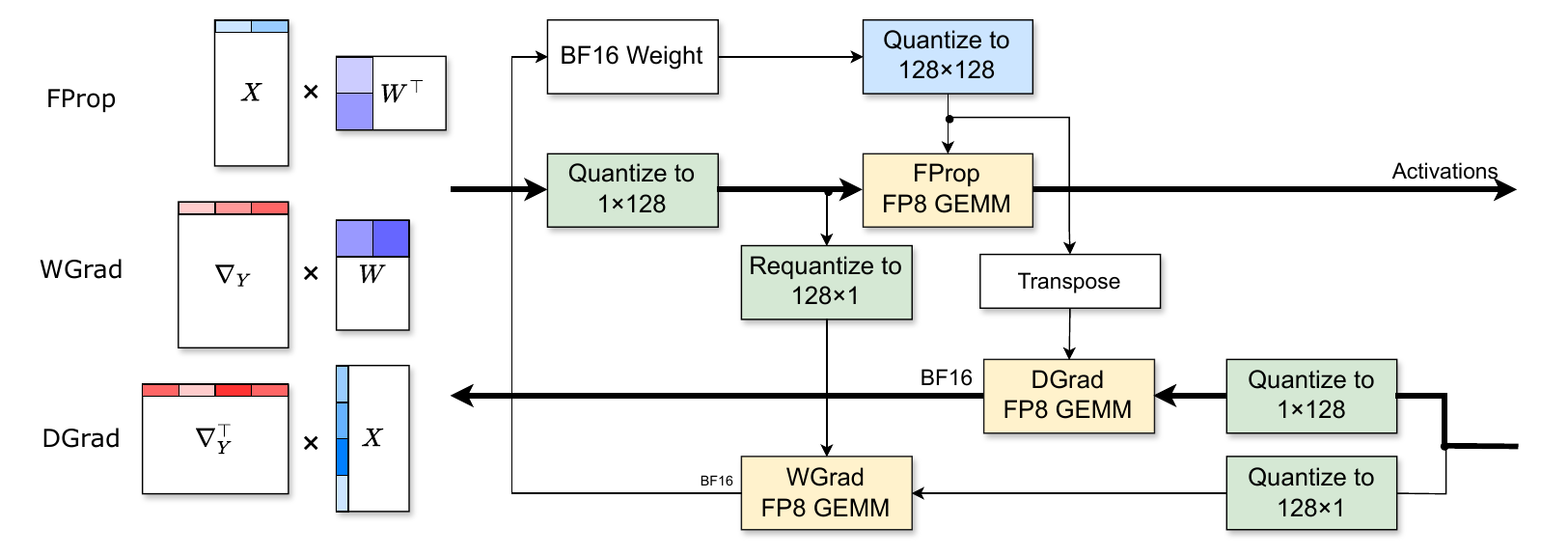}
    \caption{Our quantization scheme for a linear layer. The FProp GEMM and the WGrad GEMM uses an $(1\times 128) \times  (128 \times 128)$ FP8 Matmul kernel, while the DGrad GEMM uses an $(1\times 128) \times  (128 \times 1)$ FP8 matmul kernel. }
    \lblfig{linear_layer_granularity}
\end{figure*}


We find that the root cause of \naivefpeight{} training failure is the \textbf{inconsistency of the precision flow} between training and rollout, making the RL training effectively off-policy. Given the critical importance of maintaining on-policy consistency in RL~\cite{liu2025flashrl}, addressing this issue is essential.
To mitigate this issue, we propose enforcing an \textbf{unified FP8 precision flow} between training and rollout, ensuring that the responses in rollout are consistent with training, making it an on-policy process.

Formally, we model the propagation of quantization precision across the model as a directed graph $\gG = (\gV, \gE)$, as illustrated in \fig{quantization_precision_flow}. A node $v_i \in \mathcal{V}$ represents either an operator or a weight within the model, which we refer to as operator nodes or weight nodes. A directed edge $(v, v^\prime) \in \mathcal{E}$ is connected when the output of $v$ serves as the input of $v^\prime$ to describe tensor propagation between two connected operators. An edge indicates the precision of the transferred tensor and the quantization granularity of the tensor (if quantized).

In training, we have the graph $\gG_{\text{train}}$, which can be further separated into two subgraphs $\gG_{\text{train}}^{\text{fwd}}$ and  $\gG_{\text{train}}^{\text{bwd}}$. Within $\gG_{\text{train}}^{\text{fwd}}$, edges represent activations, while in $\gG_{\text{train}}^{\text{bwd}}$, edges represent gradients. They share the same nodes and topology since they have the same suite of operators, but the edge direction is reversed. These two graphs are connected by edges representing activations saved for the backward pass.

For the inference engine, we have the graph $\gG_{\text{infer}}$, which also has the same topology as $\gG_{\text{train}}^{\text{fwd}}$ when using BF16 inference. In the FP8 rollout scenario, the weight nodes and their edges are in FP8 precision.

For \naivefpeight{}, all weights and activations in the training graph are in BF16. However, in its inference graph $\gG_{\text{infer}}$, all edges that feed into a linear layer are quantized into FP8.
\textbf{This makes \naivefpeight{} have two distinct forward quantization graphs: $\gG^{\text{fwd}}_{\text{train}}$ and $\gG_{\text{infer}}$.} This leads to a mismatch between training and rollout, making the forward process deviate from what the actor will actually generate during rollout. Therefore, the RL training is off-policy, unstable, and unlikely to achieve satisfactory results.

\subsection{Unified FP8 Precision Flow between Training and Rollout}

In \method{}, we propose solving this problem by forcing $\gG_{\text{infer}}$ to be a subgraph of $\gG_{\text{train}}^{\text{fwd}}$. 
All other attributes (precision and granularity) of the edges are kept the same. The only difference is that $\gG_{\text{train}}^{\text{fwd}}$ has a higher-precision master copy to stabilize training, since the master weight of $\gG_{\text{train}}$ needs to be stored in BF16. We ensure that the forward pass of training and inference frameworks shares consistent quantization behavior, thereby mitigating the mismatch in precision propagation observed in previous approaches. This is demonstrated in \fig{quantization_precision_flow}.

For $\gG_{\text{train}}^{\text{bwd}}$, we first focus on the activations saved for backward computation. Consider an edge $(v, v^\prime)$ in the forward pass, where the tensor is in FP8 precision and $v^\prime$ is a GEMM operator. The $v^\prime$ operator can only access the FP8 tensor, since the quantization is usually fused with the previous step. Therefore, we elect to store the activations for the backward pass also in FP8 precision. This strategy has been shown to maintain training stability in prior work on pretraining and supervised fine-tuning~\cite{xi2024coat}. 

We retain the gradients transported between operators during the backward pass in BF16 precision to preserve model accuracy. Although quantizing them can further reduce communication overhead, they often introduce gradient underflow or quantization noise that degrades convergence.
The GEMMs in the backward pass (DGrad and WGrad) are also quantized into FP8 precision for acceleration. We elaborate on the details in the following section.

\subsection{Granularity of GEMM Quantization}

We quantize all GEMM operators in training and inference for the linear layers to accelerate computation. As shown in \fig{linear_layer_granularity}, the FProp operator in the forward pass and the WGrad and DGrad operators in the backward pass all take FP8 tensors as input and output BF16 tensors. In this section, we discuss the quantization scheme adopted for these FP8 GEMMs.

Per-tensor quantization of FP8 has been shown to be unstable in training large language models. Therefore, we adopt a finer-grained quantization granularity when quantizing the activations, weights, and gradients into FP8 precision. Specifically, we quantize the weights using $128\times 128$ per-block quantization and quantize the activations and gradients using $1\times 128$ per-group quantization. We now discuss the strategy for each GEMM below.

\paragraph{FProp Operator}
For the FProp operator, the input activation is quantized with $1\times 128$ per-group quantization, while the weight is quantized with $128\times 128$ per-block quantization. As described in \sect{quantization_linear_layer}, the hardware kernel needs to be row-wise $\times$ column-wise, so both the activation and weight are stored in a row-wise layout. The quantization of activation can be fused with its previous operator to reduce overhead, while the quantization of weight needs to be done explicitly during training. For weight quantization in inference, we quantize the weight during the parameter update stage, which incurs negligible overhead and is fully compatible with tensor parallelism.

\paragraph{DGrad Operator and WGrad Operator}
For the DGrad operator, its workload is equivalent to the FProp operator, which is a $1\times 128$ quantized matrix multiplies a $128\times 128$ quantized matrix. Thus, it can directly reuse the same kernel configuration.
For the WGrad operator, following DeepSeek-V3, we quantize the first matrix in $1\times 128$ while quantizing the second one in $128\times 1$. This finer-grained design helps stabilize training.

These two operators both require the quantized gradient, but one requires it to be $1\times 128$ quantized while the other requires it to be $128 \times 1$ quantized, so we fuse these quantization processes. Since the weight's quantization scheme is symmetric along the channel and row axis, its value does not change in the backward pass, so we only need to perform a transpose in the backward pass. For the activations, in the forward pass they are quantized as $1\times 128$, but in the backward pass, they need to be quantized as $128\times 1$. This discrepancy forces us to quantize them again in the backward pass. 
This is even beneficial for quantized training.

\subsection{Implementation}
We use vLLM~\cite{Kwon2023EfficientMM} to serve as the inference engine and VeRL~\cite{sheng2024hybridflow_verl} as our RL training framework. For the quantized GEMMs, we reference the kernels from DeepGEMM~\cite{deepseek2025deepgemm}. We use Triton~\cite{tillet2019triton} to implement the quantization, transpose, and fused activation or RMSNorm kernels.
\section{Evaluation}

{
\renewcommand{\arraystretch}{1.2}
\begin{table*}[t]
\centering
\small
\begin{tabular}{ll|llll|l}

\toprule
Model & Training Precision & GSM8k & MATH 500 & GPQA & SuperGPQA & Average$\uparrow$ \\
\midrule

\multicolumn{2}{l}{\textit{Llama3.1-8B}} & \multicolumn{2}{l}{\textit{Rollout Length = 8k}} & \multicolumn{2}{l}{\textit{Dataset = GSM8k + MATH}} \\
\cmidrule(lr){1-7}
 & Initial       & 0.8 & 3.3 & 17.3 & 12.4 & 8.5 \\
\cmidrule(lr){2-7}
 & BF16                & 49.0 & 12.1 & 15.7 & 15.9 & 23.2 \\
 & \NaiveFpEight   & 20.6\loss{28.4} & 5.4\loss{6.7} & 11.2\loss{4.5} & 14.7\loss{1.2} & 13.0\loss{10.2} \\
 \rowcolor{gray!10}
 & \method{}           & 47.2\loss{1.8} & 11.3\loss{0.8} & 22.3\gain{6.6} & 19.9\gain{4.0} & 25.2\gain{2.0} \\
\cmidrule(lr){1-7}

\multicolumn{2}{l}{\textit{Qwen2.5-7B}} & \multicolumn{2}{l}{\textit{Rollout Length = 8k}} & \multicolumn{2}{l}{\textit{Dataset = GSM8k + MATH}} \\
\cmidrule(lr){1-7}
 & Initial       & 17.5 & 46.5 & 28.9 & 25.5 & 29.6 \\
\cmidrule(lr){2-7}
 & BF16                & 91.8 & 71.0 & 36.0 & 28.6 & 56.9 \\
 & \NaiveFpEight   & \multicolumn{5}{c}{\textit{Did not converge}} \\
 \rowcolor{gray!10}
 & \method{}           & 89.9\loss{1.9} & 69.5\loss{1.5} & 35.5\loss{0.5} & 28.5\loss{0.1} & 55.9\loss{1.0} \\
\cmidrule(lr){1-7}

\multicolumn{2}{l}{\textit{Qwen3-8B-Base}} & \multicolumn{2}{l}{\textit{Rollout Length = 8k}} & \multicolumn{2}{l}{\textit{Dataset = GSM8k + MATH}} \\
\cmidrule(lr){1-7}
 & Initial        & 63.4 & 69.7 & 46.2 & 31.8 & 52.6 \\
\cmidrule(lr){2-7}
 & BF16                & 92.9 & 83.3 & 43.7 & 35.1 & 63.8 \\
 & \NaiveFpEight   & 92.1\loss{0.8} & 77.0\loss{6.3} & 44.2\gain{0.5} & 30.3\loss{4.8} & 60.9\loss{2.9} \\
 \rowcolor{gray!10}
 & \method{}           & 93.1\gain{1.0} & 81.3\loss{2.0} & 42.1\loss{1.6} & 35.2\gain{0.1} & 62.7\loss{1.1} \\
 \bottomrule
\end{tabular}
\caption{\textbf{Performance comparison of \method{} and baselines when rollout length is set to 8k.} In all settings of models, \method{} greatly reduces the degradation introduced by \naivefpeight{} and achieves close performance to BF16 RL training. 
Green (red) micro-numbers indicate absolute gain (drop) vs.\ \emph{BF16} within each model block; blank means the baseline did not converge.}
\lbltbl{main_table_8k}
\end{table*}
{
\renewcommand{\arraystretch}{1.2}
\begin{table*}[t]
\centering
\small

\begin{tabular}{ll|llll|l}
\toprule
Model & Training Precision & GSM8k & MATH 500 & GPQA & SuperGPQA & Average$\uparrow$ \\
\midrule

\multicolumn{2}{l}{\textit{Qwen2.5-7B}} & \multicolumn{2}{l}{\textit{Rollout Length = 16k}} & \multicolumn{2}{l}{\textit{Dataset = GSM8k + MATH}} \\
\cmidrule(lr){1-7}
 & Before Tuning       & 13.5 & 45.3 & 28.4 & 26.0 & 28.3 \\
\cmidrule(lr){2-7}
 & BF16                & 91.5 & 72.8 & 28.4 & 42.2 & 58.7 \\
 & \NaiveFpEight   & 90.7\loss{0.8} & 68.1\loss{4.7} & 28.9\gain{0.5} & 26.9\loss{15.3} & 53.7\loss{5.0} \\
 \rowcolor{gray!10}
 & \method{}           & 89.3\loss{2.2} & 69.2\loss{3.6} & 36.5\gain{8.1} & 27.8\loss{14.4} & 55.7\loss{3.0} \\
\cmidrule(lr){1-7}

\multicolumn{2}{l}{\textit{Qwen3-8B-Base}} & \multicolumn{2}{l}{\textit{Rollout Length = 16k}} & \multicolumn{2}{l}{\textit{Dataset = GSM8k + MATH}} \\
\cmidrule(lr){1-7}
 & Before Tuning       & 63.7 & 69.7 & 45.7 & 30.7 & 52.5\\
\cmidrule(lr){2-7}
 & BF16                & 94.7 & 81.7 & 46.2 & 33.5 & 64.0 \\
 & \NaiveFpEight   & \multicolumn{5}{c}{\textit{Did not converge}} \\
 \rowcolor{gray!10}
 & \method{}           & {92.1}\loss{2.6} & {76.7}\loss{5.0} & {43.3}\loss{2.9} & {33.2}\loss{0.3} & {61.3}\loss{2.7} \\
\cmidrule(lr){1-7}
 
\multicolumn{2}{l}{\textit{Qwen3-8B-Base}} & \multicolumn{2}{l}{\textit{Rollout Length = 16k}} & \multicolumn{2}{l}{\textit{Dataset = DeepMATH}} \\
\cmidrule(lr){1-7}
 & Before Tuning       & 63.4 & 69.7 & 46.2 & 31.8 & 52.8 \\
\cmidrule(lr){2-7}
 & BF16                & * & 83.4 & 44.2 & 36.1 & 54.6 \\
 & \NaiveFpEight   & * & 57.8\loss{25.6} & 42.6\loss{1.6} & 32.6\loss{3.5} & 44.3\loss{10.3} \\
 \rowcolor{gray!10}
 & \method{}           & * & {80.2}\loss{3.2} & {47.2}\gain{3.0} & {33.8}\loss{2.3} & {53.7}\loss{0.9} \\

\bottomrule
\end{tabular}

\caption{\textbf{Performance comparison of \method{} and baseline when rollout length is 16k, or trained on DeepMATH}. In all settings of models and dataset, \method{} greatly reduce the degradation introduced by \naivefpeight{} and achieve close performance to BF16 RL training.}
\lbltbl{main_table_16k}

\end{table*}

}

\subsection{Evaluation Setup}

 \paragraph{Models.} We evaluate \method{} on several widely used open-source models, including Llama3.1-8B, Qwen2.5-7B, and Qwen3-8B-Base. For Llama3.1-8B and Qwen2.5-7B, we set the rollout length to 8K. For Qwen3-8B-Base, we train on both 8K and 16K rollout lengths.

\paragraph{Datasets and RL Training Setting.} We conduct experiments under two dataset settings. We first train on a mixture of GSM8K and MATH, where we set the rollout generation number to 4. GSM8K contains 8,500 grade school math word problems, and MATH contains 12,500 complex math competition problems.
We then train on the DeepMATH dataset~\cite{he2025deepmath103klargescalechallengingdecontaminated}, where we set the rollout generation number to 16. DeepMATH contains 103K math problems designed with high difficulty.

\paragraph{Hyperparameters.} For all experiments, we set the learning rate to $10^{-6}$ and the batch size to 256. The KL loss coefficient is set to $10^{-3}$. The evaluation of checkpoints is conducted every 5 steps during RL training. We conduct the experiments on NVIDIA H100 GPUs. 

\paragraph{Evaluation Metrics.}

We evaluate the final checkpoints on 5 downstream benchmarks: GSM8K~\cite{Cobbe2021GSM8k}, MATH500~\cite{hendrycks2021measuringmathematicalproblemsolving}, AMC, GPQA~\cite{rein2023gpqagraduatelevelgoogleproofqa}, and SuperGPQA~\cite{pteam2025supergpqascalingllmevaluation}, and their average score. We evaluate GSM8K and MATH500 on their test splits. For AMC, GPQA, and SuperGPQA, we follow the preprocessing strategy of Guru-92k~\cite{cheng2025Reasoning360}.

\subsection{Accuracy Evaluation}
We first analyze the results when using an 8K rollout length. As shown in \tbl{main_table_8k}, which evaluates models using an 8K rollout length on the GSM8K + MATH dataset, the \naivefpeight{} method exhibits significant instability. Most notably, it \textbf{fails to converge} entirely on the Qwen2.5-7B model. On models where it does converge, it incurs substantial performance degradation compared to the BF16 baseline. For instance, on Llama3.1-8B, the average score drops by 9.8\%, and on Qwen3-8B-Base, it drops by 2.9\%. In sharp contrast, our \method{} method proves robust and effective. It not only converges in all scenarios but also greatly reduces the gap to BF16 training. On Llama3.1-8B, \method{} even outperforms the BF16 baseline by 2.0\%. On Qwen2.5-7B, the performance degradation is merely 1.0\% (56.9\% vs 55.9\%), and on Qwen3-8B-Base it degrades by only 1.1\% (63.8\% vs 62.7\%).

We analyze the performance under more challenging configurations, including a 16K rollout length and the DeepMATH dataset, as presented in \tbl{main_table_16k}. This further underscores the instability of the \naivefpeight{} method, particularly under more challenging configurations such as a 16K rollout length or a different training dataset (DeepMATH). The \naivefpeight{} method \textbf{fails to converge} on the Qwen3-8B-Base model with a 16K rollout. On the Qwen3-8B-Base (DeepMATH) experiment, it suffers a severe performance degradation of 10.3\% compared to the BF16 baseline. On the Qwen2.5-7B (16K) model, it also shows a notable degradation of 5.0\%. In contrast, \method{} successfully resolves these issues. It converges on the Qwen3-8B model, closing the gap to just 2.7\%. On the Qwen3-8B-Base (DeepMATH) experiment, \method{} reduce the gap to 0.9\% (54.6\% vs 53.7\%). On the Qwen2.5-7B model, the degradation is also reduced from 5.0\% to 3.0\%.
The results demonstrate that \method{} framework achieves more stable convergence and better performance compared to \naivefpeight{}, and \method{} is closely aligned with the BF16 baselines.


\subsection{Efficiency Evaluation}

\begin{table}[t]
\centering
\small
\setlength{\tabcolsep}{6pt}
\renewcommand{\arraystretch}{1.2}
\begin{tabular}{lrrr}
\toprule
\textbf{Model \& TP} & \textbf{4K} & \textbf{8K} & \textbf{16K} \\
\midrule
8B (TP=1)  & 1.12$\times$ & 1.10$\times$ & 1.12$\times$ \\
14B (TP=1) & 1.26$\times$ & 1.28$\times$ & 1.29$\times$ \\
14B (TP=2) & 1.08$\times$ & 1.12$\times$ & 1.10$\times$ \\
32B (TP=2) & 1.29$\times$ & 1.33$\times$ & 1.30$\times$ \\
32B (TP=4) & 1.08$\times$ & 1.07$\times$ & 1.10$\times$ \\
\bottomrule
\end{tabular}
\caption{\textbf{Rollout speedup in FP8 over BF16 in tokens/s at different output lengths.} Used 512 prompts with max concurrent request 128, and 512 input length. Measured on H100s.}
\lbltbl{fp8_speedup}
\end{table}

We quantify rollout efficiency gains from FP8 with an offline generation benchmark in vLLM~\cite{Kwon2023EfficientMM}, reporting throughput speedup from FP8. We vary the model size from 8B to 32B, and test under multiple tensor parallel settings.
As summarized in \tbl{fp8_speedup}, FP8 achieves consistent acceleration, with speedups ranging from 1.07× to 1.33× over BF16. 

We observe two trends in the speedup ratio. First, the benefit of FP8 quantization increases with model size. Larger models, such as the 32B configuration, achieve the most substantial speedups up to 1.33, as it is more compute-intensive. This allows the optimized FP8 tensor-core kernels (e.g., DeepGEMM) to more effectively accelerate computation. In smaller models (e.g., 8B), memory access overhead constitutes a larger fraction of latency, limiting the overall benefit of FP8 inference.

Second, higher degrees of tensor parallelism (TP) decrease the observed speedup. When the model is distributed across more GPUs, the communication overhead becomes more pronounced. For example, the 32B model with TP=4 shows only $1.1×$ improvement compared to $1.3×$ at TP=2. This trend highlights that FP8 rollout can be accelerated under a lower tensor parallel degree.

For end-to-end RL training on Qwen3-8B with an 8K rollout length, FP8 quantization delivers consistent acceleration across multiple computation stages. Specifically, FP8 achieves a $1.54\times$ speedup in the actor update phase and a $1.80\times$ speedup in the reference model inference, which together contribute to an overall $1.41\times$ improvement in the training phase throughput. Combined with the speedup from rollout, this results in an end-to-end step-time speedup of $1.16\times$ for 8B model training. We expect the speedup to be much more significant for larger model sizes. A full scaling study on 14–32B models is left to future work given resource constraints.

\section{Related Works}
\footnotetext{Due to format issue we do not report GSM8k results when training on DeepMATH.}

\subsection{Low-Precision Training and Inference for LLMs}
The exponential growth in the parameter counts of Large Language Models has rendered full-precision training and inference prohibitively expensive, creating significant barriers related to memory, computation, and energy consumption~\cite{gu2025jet,dong2024hymba,blakeman2025nemotron}. This fundamental challenge has catalyzed the development of low-precision techniques, which reduce the numerical precision of model weights, activations, and gradients to lower-precision formats to leverage hardware accelerators like NVIDIA's Tensor Cores that offer significantly higher throughput for low-precision arithmetic. 

Post-Training Quantization (PTQ), which compresses a pre-trained model in a "one-shot" fashion without fine-tuning, has become the dominant approach for deploying LLMs. SmoothQuant~\cite{Xiao2022SmoothQuantAA} addresses the activation outlier problem by migrating the quantization difficulty from activations to weights. GPTQ compresses 175-billion-parameter models down to 3 or 4 bits with minimal impact on perplexity. AWQ protects salient weights by applying a per-channel scaling factor in an activation-aware, reconstruction-free approach.

On the other hand, Fully Quantized Training (FQT) accelerates the training stage of LLMs by performing computations in low precision. SwitchBack and Jetfire~\cite{xi2024jetfire}propose advanced INT8 training for transformers through an INT8 precision flow to optimize memory access and a per-block quantization method to maintain accuracy. NVIDIA Transformer Engine~\cite{nvidia_transformer_engine} accelerates transformer model training using FP8. Building on this, frameworks like COAT extend FP8 quantization beyond just linear layers to include optimizer states and activations. Concurrently, methods like QLoRA have focused on memory efficient fine-tuning using 4-bit quantization of a frozen pre-trained model with Low-Rank Adaptation (LoRA).

\subsection{Reinforcement Learning for Large Language Models}

Reinforcement Learning (RL) has played a central role in the evolution of large language models (LLMs), progressing from alignment to reasoning. Early work on alignment employed Reinforcement Learning from Human Feedback (RLHF), where a reward model trained on human preference data guided policy optimization via algorithms like Proximal Policy Optimization (PPO). Subsequent methods, such as Direct Preference Optimization (DPO), removed the explicit reward model, enabling simpler and more stable preference-based fine-tuning.   

Recent efforts have shifted toward reasoning-oriented RL, where the correctness of a model's output can be automatically verified. This shift has spurred the development of new RL algorithms tailored for reasoning tasks. For instance, the {DeepSeek-R1} model was developed using {GRPO} (Group Relative Policy Optimization), a critic-free RL algorithm that estimates advantages by comparing a response's reward to the average reward of a group of responses generated from the same prompt. GSPO was introduced to enhance training stability by performing optimization at the sequence level. DAPO introduces several techniques, including "Clip-Higher" to avoid entropy collapse and "Dynamic Sampling" to filter out uninformative training samples for better stability and efficiency.

\subsection{Efficient Reasoning with Large Language Models}

While RLVR is a powerful paradigm for training reasoning models, it is computationally intensive due to the need to generate long autoregressive rollouts and the on-policy nature of RL, which can lead to low hardware utilization. This has motivated research into both system-level and algorithmic optimizations to make reasoning more efficient.   

From a systems perspective, async RL frameworks such as AReaL have been proposed to break the synchronous dependency between rollout and training. These asynchronous systems allow rollout workers to continuously generate new data while training workers update the model, thereby improving GPU utilization and significantly increasing training throughput. ReaLHF focuses on improving the parallelism strategy of RL training. By automating the parallelization search period, ReaLHF improves GPU utilization and reduces GPU idle time.

From an algorithmic perspective, a key challenge is the phenomenon of "overthinking," where models generate excessively long and redundant reasoning trajectories. NoThinking~\cite{ma2025reasoningmodelseffectivethinking} propose to prune the reasoning trajectory and teaches models to self-regulate their reasoning process by identifying redundant steps and generating training signals that encourage earlier termination.

Finally, co-design offers another angle to tackle these efficiency challenges. The QeRL framework~\cite{huang2025qerlefficiencyquantizationenhanced} addresses the high cost of RL rollouts by combining NVFP4 quantization with Low-Rank Adaptation (LoRA). This synergy drastically reduces the memory footprint and accelerates the generation phase, enabling efficient RL training of a 32B model on a single GPU. Truncated Importance Sampling (TIS)~\cite{liu2025flashrl} proposes to mitigate the off-policy issue by adding the importance ratio to the model update and truncating it if the inference probability is small.

\section{Conclusion}

In this work, we addressed the critical performance bottleneck of the rollout phase in RL training. We demonstrated that the naive \textit{BF16-train + FP8-rollout} strategy is fundamentally flawed, as it introduces a training-rollout mismatch that leads to training instability and catastrophic performance collapse. To solve this, we proposed \method{}, a framework that enables robust on-policy FP8 RL training by adopting an identical FP8 precision flow for both training forward pass and inference rollout stage. Our comprehensive evaluations show that \method{} robustly converges across all models and benchmarks settings. Our method also maintains a competitive performance close to the BF16 RL baseline, usually less than $1\%$ degradation. By delivering up to $1.33\times$ rollout phase speedup, up to $1.41\times$ training phase speedup, and an $1.16\times$ end-to-end speedup without sacrificing model accuracy, \method{} establishes a reliable and efficient path forward for applying FP8 computation to accelerate large-scale RL training.

\clearpage

\bibliographystyle{unsrt}
\bibliography{example_paper}

@article{hu2024openrlhf,
  title={OpenRLHF: An Easy-to-use, Scalable and High-performance RLHF Framework},
  author={Jian Hu and Xibin Wu and Zilin Zhu and Xianyu and Weixun Wang and Dehao Zhang and Yu Cao},
  journal={arXiv preprint arXiv:2405.11143},
  year={2024}
}

@misc{slime_github,
      author = {Zilin Zhu and Chengxing Xie and Xin Lv and slime Contributors},
      title = {slime: An LLM post-training framework for RL Scaling},
      year = {2025},
      howpublished = {\url{https://github.com/THUDM/slime}},
      note = {GitHub repository. Corresponding author: Xin Lv}
}

@misc{nemo-rl,
    title = {NeMo RL: A Scalable and Efficient Post-Training Library},
    howpublished = {\url{https://github.com/NVIDIA-NeMo/RL}},
    year = {2025},
    note = {GitHub repository},
}

@article{sheng2024hybridflow_verl,
  title   = {HybridFlow: A Flexible and Efficient RLHF Framework},
  author  = {Guangming Sheng and Chi Zhang and Zilingfeng Ye and Xibin Wu and Wang Zhang and Ru Zhang and Yanghua Peng and Haibin Lin and Chuan Wu},
  year    = {2024},
  journal = {arXiv preprint arXiv: 2409.19256}
}

@article{jaech2024openai,
  title={Openai o1 system card},
  author={Jaech, Aaron and Kalai, Adam and Lerer, Adam and Richardson, Adam and El-Kishky, Ahmed and Low, Aiden and Helyar, Alec and Madry, Aleksander and Beutel, Alex and Carney, Alex and others},
  journal={arXiv preprint arXiv:2412.16720},
  year={2024}
}

@article{guo2025deepseek,
  title={Deepseek-r1: Incentivizing reasoning capability in llms via reinforcement learning},
  author={Guo, Daya and Yang, Dejian and Zhang, Haowei and Song, Junxiao and Zhang, Ruoyu and Xu, Runxin and Zhu, Qihao and Ma, Shirong and Wang, Peiyi and Bi, Xiao and others},
  journal={arXiv preprint arXiv:2501.12948},
  year={2025}
}

@Article{Wei2022Chainofthought,
 author = {Jason Wei and Xuezhi Wang and Dale Schuurmans and Maarten Bosma and Ed H. Chi and F. Xia and Quoc Le and Denny Zhou},
 booktitle = {Neural Information Processing Systems},
 journal = {ArXiv},
 title = {Chain of Thought Prompting Elicits Reasoning in Large Language Models},
 volume = {abs/2201.11903},
 year = {2022}
}

@misc{zhou2024chainofthoughtreasoningwithout,
      title={Chain-of-Thought Reasoning Without Prompting}, 
      author={Denny Zhou and Xuezhi Wang},
      year={2024},
      eprint={2402.10200},
      archivePrefix={arXiv},
      primaryClass={cs.CL},
      url={https://arxiv.org/abs/2402.10200}, 
}

@Article{Ouyang2022TrainingRLHF,
 author = {Long Ouyang and Jeff Wu and Xu Jiang and Diogo Almeida and Carroll L. Wainwright and Pamela Mishkin and Chong Zhang and Sandhini Agarwal and Katarina Slama and Alex Ray and John Schulman and Jacob Hilton and Fraser Kelton and Luke E. Miller and Maddie Simens and Amanda Askell and P. Welinder and P. Christiano and J. Leike and Ryan J. Lowe},
 booktitle = {Neural Information Processing Systems},
 journal = {ArXiv},
 title = {Training language models to follow instructions with human feedback},
 volume = {abs/2203.02155},
 year = {2022}
}

@Article{Bai2022TrainingRLHF2,
 author = {Yuntao Bai and Andy Jones and Kamal Ndousse and Amanda Askell and Anna Chen and Nova Dassarma and Dawn Drain and Stanislav Fort and Deep Ganguli and T. Henighan and Nicholas Joseph and Saurav Kadavath and John Kernion and Tom Conerly and S. El-Showk and Nelson Elhage and Zac Hatfield-Dodds and Danny Hernandez and Tristan Hume and Scott Johnston and Shauna Kravec and Liane Lovitt and Neel Nanda and Catherine Olsson and Dario Amodei and Tom B. Brown and Jack Clark and Sam McCandlish and C. Olah and Benjamin Mann and Jared Kaplan},
 booktitle = {arXiv.org},
 journal = {ArXiv},
 title = {Training a Helpful and Harmless Assistant with Reinforcement Learning from Human Feedback},
 volume = {abs/2204.05862},
 year = {2022}
}

@misc{yu2025dapoopensourcellmreinforcement,
      title={DAPO: An Open-Source LLM Reinforcement Learning System at Scale}, 
      author={Qiying Yu and Zheng Zhang and Ruofei Zhu and Yufeng Yuan and Xiaochen Zuo and Yu Yue and Weinan Dai and Tiantian Fan and Gaohong Liu and Lingjun Liu and Xin Liu and Haibin Lin and Zhiqi Lin and Bole Ma and Guangming Sheng and Yuxuan Tong and Chi Zhang and Mofan Zhang and Wang Zhang and Hang Zhu and Jinhua Zhu and Jiaze Chen and Jiangjie Chen and Chengyi Wang and Hongli Yu and Yuxuan Song and Xiangpeng Wei and Hao Zhou and Jingjing Liu and Wei-Ying Ma and Ya-Qin Zhang and Lin Yan and Mu Qiao and Yonghui Wu and Mingxuan Wang},
      year={2025},
      eprint={2503.14476},
      archivePrefix={arXiv},
      primaryClass={cs.LG},
      url={https://arxiv.org/abs/2503.14476}, 
}

@Inproceedings{Jin2025SearchR1TL,
 author = {Bowen Jin and Hansi Zeng and Zhenrui Yue and Dong Wang and Hamed Zamani and Jiawei Han},
 title = {Search-R1: Training LLMs to Reason and Leverage Search Engines with Reinforcement Learning},
 year = {2025}
}

@misc{cheng2025Reasoning360,
  title         = {Revisiting Reinforcement Learning for LLM Reasoning from A Cross-Domain Perspective},
  author        = {Zhoujun Cheng and Shibo Hao and Tianyang Liu and Fan Zhou and Yutao Xie and Feng Yao and Yuexin Bian and Yonghao Zhuang and Nilabjo Dey and Yuheng Zha and Yi Gu and Kun Zhou and Yuqi Wang and Yuan Li and Richard Fan and Jianshu She and Chengqian Gao and Abulhair Saparov and Haonan Li and Taylor W. Killian and Mikhail Yurochkin and Zhengzhong Liu and Eric P. Xing and Zhiting Hu},
  journal       = {arXiv preprint arXiv:2506.14965},
  year          = {2025},
  doi           = {10.48550/arXiv.2506.14965},
  url           = {https://arxiv.org/abs/2506.14965}
}

@article{chollet2024arc,
  title={Arc prize 2024: Technical report},
  author={Chollet, Francois and Knoop, Mike and Kamradt, Gregory and Landers, Bryan},
  journal={arXiv preprint arXiv:2412.04604},
  year={2024}
}

@misc{chollet2025arcagi2newchallengefrontier,
      title={ARC-AGI-2: A New Challenge for Frontier AI Reasoning Systems}, 
      author={Francois Chollet and Mike Knoop and Gregory Kamradt and Bryan Landers and Henry Pinkard},
      year={2025},
      eprint={2505.11831},
      archivePrefix={arXiv},
      primaryClass={cs.AI},
      url={https://arxiv.org/abs/2505.11831}, 
}

@misc{liu2025flashrl,
  title = {FlashRL: 8Bit Rollouts, Full Power RL},
  url = {https://fengyao.notion.site/flash-rl},
  author = {Liu, Liyuan and Yao, Feng and Zhang, Dinghuai and Dong, Chengyu and Shang, Jingbo and Gao, Jianfeng},
  journal = {Feng Yao's Notion},
  year = {2025},
  month = aug,
}

@Article{Lin2023AWQ,
 author = {Ji Lin and Jiaming Tang and Haotian Tang and Shang Yang and Xingyu Dang and Song Han},
 booktitle = {Conference on Machine Learning and Systems},
 journal = {GetMobile Mob. Comput. Commun.},
 pages = {12-17},
 title = {AWQ: Activation-aware Weight Quantization for On-Device LLM Compression and Acceleration},
 volume = {28},
 year = {2023}
}

@Article{Frantar2022GPTQ,
 author = {Elias Frantar and Saleh Ashkboos and T. Hoefler and Dan Alistarh},
 booktitle = {arXiv.org},
 journal = {ArXiv},
 title = {GPTQ: Accurate Post-Training Quantization for Generative Pre-trained Transformers},
 volume = {abs/2210.17323},
 year = {2022}
}

@Article{Xiao2022SmoothQuantAA,
 author = {Guangxuan Xiao and Ji Lin and Mickael Seznec and Julien Demouth and Song Han},
 booktitle = {International Conference on Machine Learning},
 journal = {ArXiv},
 title = {SmoothQuant: Accurate and Efficient Post-Training Quantization for Large Language Models},
 volume = {abs/2211.10438},
 year = {2022}
}

@misc{yao2025offpolicy,
  title = {Your Efficient RL Framework Secretly Brings You Off-Policy RL Training},
  url = {https://fengyao.notion.site/off-policy-rl},
  author = {Yao, Feng and Liu, Liyuan and Zhang, Dinghuai and Dong, Chengyu and Shang, Jingbo and Gao, Jianfeng},
  journal = {Feng Yao's Notion},
  year = {2025},
  month = aug,
}

@misc{zheng2025prosperitycollapsefaroffpolicy,
      title={Prosperity before Collapse: How Far Can Off-Policy RL Reach with Stale Data on LLMs?}, 
      author={Haizhong Zheng and Jiawei Zhao and Bedi Chen},
      year={2025},
      eprint={2510.01161},
      archivePrefix={arXiv},
      primaryClass={cs.LG},
      url={https://arxiv.org/abs/2510.01161}, 
}

@misc{cetin2022stabilizingoffpolicydeepreinforcement,
      title={Stabilizing Off-Policy Deep Reinforcement Learning from Pixels}, 
      author={Edoardo Cetin and Philip J. Ball and Steve Roberts and Oya Celiktutan},
      year={2022},
      eprint={2207.00986},
      archivePrefix={arXiv},
      primaryClass={cs.LG},
      url={https://arxiv.org/abs/2207.00986}, 
}

@misc{cheng2025deepseekv3technicalreport,
      title={DeepSeek-V3 Technical Report}, 
      author={Xin Cheng and Xiaodong Liu and Yanping Huang and Zhengyan Zhang and Peng Zhang and Jiashi Li and Xinyu Yang and Damai Dai and Hui Li and Yao Zhao and Yu Wu and Chengqi Deng and Liang Zhao and H. Zhang and Kexin Huang and Junlong Li and Yang Zhang and Lei Xu and Zhen Zhang and Meng Li and Kai Hu and DeepSeek-AI and Qihao Zhu and Daya Guo and Zhihong Shao and Dejian Yang and Peiyi Wang and Runxin Xu and Huazuo Gao and Shirong Ma and Wangding Zeng and Xiao Bi and Zihui Gu and Hanwei Xu and Kai Dong and Liyue Zhang and Yishi Piao and Zhibin Gou and Zhenda Xie and Zhewen Hao and Bingxuan Wang and Junxiao Song and Zhen Huang and Deli Chen and Xin Xie and Kang Guan and Yuxiang You and Aixin Liu and Qiushi Du and Wenjun Gao and Qinyu Chen and Yaohui Wang and Chenggang Zhao and Chong Ruan and Fuli Luo and Wenfeng Liang and Yaohui Li and Yuxuan Liu and Xin Liu and Shiyu Wang and Jiawei Wang and Ziyang Song and Ying Tang and Yuheng Zou and Guanting Chen and Shanhuang Chen and Honghui Ding and Zhe Fu and Kaige Gao and Ruiqi Ge and Jianzhong Guo and Guangbo Hao and Ying He and Panpan Huang and Erhang Li and Guowei Li and Yao Li and Fangyun Lin and Wen Liu and Yiyuan Liu and Shanghao Lu and Xiaotao Nie and Tian Pei and Junjie Qiu and Hui Qu and Zehui Ren and Zhangli Sha and Xuecheng Su and Yaofeng Sun and Minghui Tang and Ziwei Xie and Yiliang Xiong and Yanhong Xu and Shuiping Yu and Xingkai Yu and Haowei Zhang and Lecong Zhang and Mingchuan Zhang and Minghua Zhang and Wentao Zhang and Yichao Zhang and Shangyan Zhou and Shunfeng Zhou and Huajian Xin and Yi Yu and Yuyang Zhou and Yi Zheng and Lean Wang and Yifan Shi and Xiaohan Wang and Wanjia Zhao and Han Bao and Wei An and Yongqiang Guo and Xiaowen Sun and Yixuan Tan and Shengfeng Ye and Yukun Zha and Xinyi Zhou and Zijun Liu and Bing Xue and Xiaokang Zhang and T. Wang and Mingming Li and Jian Liang and Jin Chen and Xiaokang Chen and Zhiyu Wu and Yiyang Ma and Xingchao Liu and Zizheng Pan and Chenyu Zhang and Yuchen Zhu and Yue Gong and Zhuoshu Li and Zhipeng Xu and Runji Wang and Haocheng Wang and Shuang Zhou and Ruoyu Zhang and Jingyang Yuan and Yisong Wang and Xiaoxiang Wang and Jingchang Chen and Xinyuan Li and Zhigang Yan and Kuai Yu and Zhongyu Zhang and Tianyu Sun and Yuting Yan and Yunfan Xiong and Yuxiang Luo and Ruisong Zhang and X.Q. Li and Zhicheng Ma and Bei Feng and Dongjie Ji and J.L. Cai and Jiaqi Ni and Leyi Xia and Miaojun Wang and Ning Tian and R.J. Chen and R.L. Jin and Ruizhe Pan and Ruyi Chen and S.S. Li and Shaoqing Wu and W.L. Xiao and Xiangyue Jin and Xianzu Wang and Xiaojin Shen and Xiaosha Chen and Xinnan Song and Y.K. Li and Y.X. Wei and Y.X. Zhu and Yuduan Wang and Yunxian Ma and Z.Z. Ren and Zilin Li and Ziyi Gao and Zhean Xu and Bochao Wu and Chengda Lu and Fucong Dai and Litong Wang and Qiancheng Wang and Shuting Pan and Tao Yun and Wenqin Yu and Xinxia Shan and Xuheng Lin and Y.Q. Wang and Yuan Ou and Yujia He and Z.F. Wu and Zijia Zhu and et al. (133 additional authors not shown)},
      year={2025},
      eprint={2412.19437},
      archivePrefix={arXiv},
      primaryClass={cs.CL},
      url={https://arxiv.org/abs/2412.19437}, 
}

@article{xi2024coat,
  title={COAT: Compressing Optimizer states and Activation for Memory-Efficient FP8 Training},
  author={Xi, Haocheng and Cai, Han and Zhu, Ligeng and Lu, Yao and Keutzer, Kurt and Chen, Jianfei and Han, Song},
  journal={arXiv preprint arXiv:2410.19313},
  year={2024}
}

@software{deepseek2025deepgemm,
  author       = {DeepSeek-AI},
  title        = {DeepGEMM: Clean and efficient FP8 GEMM kernels with fine-grained scaling},
  year         = {2025},
  month        = oct,
  note         = {GitHub repository},
  url          = {https://github.com/deepseek-ai/DeepGEMM}
}

@article{micikevicius2023ocp,
  title={OCP 8-bit floating point specification (OFP8)},
  author={Micikevicius, Paulius and Oberman, Stuart and Dubey, Pradeep and Cornea, Marius and Rodriguez, Andres and Bratt, Ian and Grisenthwaite, Richard and Jouppi, Norm and Chou, Chiachen and Huffman, Amber and others},
  journal={Open Compute Project},
  year={2023}
}

@article{micikevicius2022fp8,
  title={Fp8 formats for deep learning},
  author={Micikevicius, Paulius and Stosic, Dusan and Burgess, Neil and Cornea, Marius and Dubey, Pradeep and Grisenthwaite, Richard and Ha, Sangwon and Heinecke, Alexander and Judd, Patrick and Kamalu, John and others},
  journal={arXiv preprint arXiv:2209.05433},
  year={2022}
}

@article{xi2024jetfire,
  title={Jetfire: Efficient and accurate transformer pretraining with int8 data flow and per-block quantization},
  author={Xi, Haocheng and Chen, Yuxiang and Zhao, Kang and Teh, Kai Jun and Chen, Jianfei and Zhu, Jun},
  journal={arXiv preprint arXiv:2403.12422},
  year={2024}
}

@article{gu2025jet,
  title={Jet-Nemotron: Efficient Language Model with Post Neural Architecture Search},
  author={Gu, Yuxian and Hu, Qinghao and Yang, Shang and Xi, Haocheng and Chen, Junyu and Han, Song and Cai, Han},
  journal={arXiv preprint arXiv:2508.15884},
  year={2025}
}

@article{dong2024hymba,
  title={Hymba: A hybrid-head architecture for small language models},
  author={Dong, Xin and Fu, Yonggan and Diao, Shizhe and Byeon, Wonmin and Chen, Zijia and Mahabaleshwarkar, Ameya Sunil and Liu, Shih-Yang and Van Keirsbilck, Matthijs and Chen, Min-Hung and Suhara, Yoshi and others},
  journal={arXiv preprint arXiv:2411.13676},
  year={2024}
}

@article{blakeman2025nemotron,
  title={Nemotron-h: A family of accurate and efficient hybrid mamba-transformer models},
  author={Blakeman, Aaron and Basant, Aarti and Khattar, Abhinav and Renduchintala, Adithya and Bercovich, Akhiad and Ficek, Aleksander and Bjorlin, Alexis and Taghibakhshi, Ali and Deshmukh, Amala Sanjay and Mahabaleshwarkar, Ameya Sunil and others},
  journal={arXiv preprint arXiv:2504.03624},
  year={2025}
}

@misc{schulman2017proximalpolicyoptimizationalgorithms,
      title={Proximal Policy Optimization Algorithms}, 
      author={John Schulman and Filip Wolski and Prafulla Dhariwal and Alec Radford and Oleg Klimov},
      year={2017},
      eprint={1707.06347},
      archivePrefix={arXiv},
      primaryClass={cs.LG},
      url={https://arxiv.org/abs/1707.06347}, 
}

@misc{schulman2018highdimensionalcontinuouscontrolusing,
      title={High-Dimensional Continuous Control Using Generalized Advantage Estimation}, 
      author={John Schulman and Philipp Moritz and Sergey Levine and Michael Jordan and Pieter Abbeel},
      year={2018},
      eprint={1506.02438},
      archivePrefix={arXiv},
      primaryClass={cs.LG},
      url={https://arxiv.org/abs/1506.02438}, 
}

@Book{Kwon2023EfficientMM,
 author = {Woosuk Kwon and Zhuohan Li and Siyuan Zhuang and Ying Sheng and Lianmin Zheng and Cody Hao Yu and Joseph E. Gonzalez and Haotong Zhang and Ion Stoica},
 booktitle = {Symposium on Operating Systems Principles},
 journal = {Proceedings of the 29th Symposium on Operating Systems Principles},
 title = {Efficient Memory Management for Large Language Model Serving with PagedAttention},
 year = {2023}
}

@Article{Zheng2023SGLangEE,
 author = {Lianmin Zheng and Liangsheng Yin and Zhiqiang Xie and Chuyue Sun and Jeff Huang and Cody Hao Yu and Shiyi Cao and Christos Kozyrakis and Ion Stoica and Joseph E. Gonzalez and Clark W. Barrett and Ying Sheng},
 booktitle = {Neural Information Processing Systems},
 title = {SGLang: Efficient Execution of Structured Language Model Programs},
 year = {2023}
}

@Article{Zhao2023PyTorchFE,
 author = {Yanli Zhao and A. Gu and R. Varma and Liangchen Luo and Chien-chin Huang and Min Xu and Less Wright and Hamid Shojanazeri and Myle Ott and Sam Shleifer and Alban Desmaison and Can Balioglu and Bernard Nguyen and Geeta Chauhan and Y. Hao and Shen Li},
 booktitle = {Proceedings of the VLDB Endowment},
 journal = {Proc. VLDB Endow.},
 pages = {3848-3860},
 title = {PyTorch FSDP: Experiences on Scaling Fully Sharded Data Parallel},
 volume = {16},
 year = {2023}
}

@Article{Shoeybi2019MegatronLMTM,
 author = {M. Shoeybi and M. Patwary and Raul Puri and P. LeGresley and J. Casper and Bryan Catanzaro},
 booktitle = {arXiv.org},
 journal = {ArXiv},
 title = {Megatron-LM: Training Multi-Billion Parameter Language Models Using Model Parallelism},
 volume = {abs/1909.08053},
 year = {2019}
}

@inproceedings{rasley2020deepspeed,
  title={Deepspeed: System optimizations enable training deep learning models with over 100 billion parameters},
  author={Rasley, Jeff and Rajbhandari, Samyam and Ruwase, Olatunji and He, Yuxiong},
  booktitle={Proceedings of the 26th ACM SIGKDD international conference on knowledge discovery \& data mining},
  pages={3505--3506},
  year={2020}
}

@article{yang2025qwen3,
  title={Qwen3 technical report},
  author={Yang, An and Li, Anfeng and Yang, Baosong and Zhang, Beichen and Hui, Binyuan and Zheng, Bo and Yu, Bowen and Gao, Chang and Huang, Chengen and Lv, Chenxu and others},
  journal={arXiv preprint arXiv:2505.09388},
  year={2025}
}

@Article{Cobbe2021GSM8k,
 author = {K. Cobbe and Vineet Kosaraju and Mohammad Bavarian and Mark Chen and Heewoo Jun and Lukasz Kaiser and Matthias Plappert and Jerry Tworek and Jacob Hilton and Reiichiro Nakano and Christopher Hesse and John Schulman},
 booktitle = {arXiv.org},
 journal = {ArXiv},
 title = {Training Verifiers to Solve Math Word Problems},
 volume = {abs/2110.14168},
 year = {2021}
}

@misc{hendrycks2021measuringmathematicalproblemsolving,
      title={Measuring Mathematical Problem Solving With the MATH Dataset}, 
      author={Dan Hendrycks and Collin Burns and Saurav Kadavath and Akul Arora and Steven Basart and Eric Tang and Dawn Song and Jacob Steinhardt},
      year={2021},
      eprint={2103.03874},
      archivePrefix={arXiv},
      primaryClass={cs.LG},
      url={https://arxiv.org/abs/2103.03874}, 
}

@misc{qwen2025qwen25technicalreport,
      title={Qwen2.5 Technical Report}, 
      author={Qwen and : and An Yang and Baosong Yang and Beichen Zhang and Binyuan Hui and Bo Zheng and Bowen Yu and Chengyuan Li and Dayiheng Liu and Fei Huang and Haoran Wei and Huan Lin and Jian Yang and Jianhong Tu and Jianwei Zhang and Jianxin Yang and Jiaxi Yang and Jingren Zhou and Junyang Lin and Kai Dang and Keming Lu and Keqin Bao and Kexin Yang and Le Yu and Mei Li and Mingfeng Xue and Pei Zhang and Qin Zhu and Rui Men and Runji Lin and Tianhao Li and Tianyi Tang and Tingyu Xia and Xingzhang Ren and Xuancheng Ren and Yang Fan and Yang Su and Yichang Zhang and Yu Wan and Yuqiong Liu and Zeyu Cui and Zhenru Zhang and Zihan Qiu},
      year={2025},
      eprint={2412.15115},
      archivePrefix={arXiv},
      primaryClass={cs.CL},
      url={https://arxiv.org/abs/2412.15115}, 
}

@inproceedings{tillet2019triton,
  title={Triton: an intermediate language and compiler for tiled neural network computations},
  author={Tillet, Philippe and Kung, Hsiang-Tsung and Cox, David},
  booktitle={Proceedings of the 3rd ACM SIGPLAN International Workshop on Machine Learning and Programming Languages},
  pages={10--19},
  year={2019}
}

@misc{he2025deepmath103klargescalechallengingdecontaminated,
      title={DeepMath-103K: A Large-Scale, Challenging, Decontaminated, and Verifiable Mathematical Dataset for Advancing Reasoning}, 
      author={Zhiwei He and Tian Liang and Jiahao Xu and Qiuzhi Liu and Xingyu Chen and Yue Wang and Linfeng Song and Dian Yu and Zhenwen Liang and Wenxuan Wang and Zhuosheng Zhang and Rui Wang and Zhaopeng Tu and Haitao Mi and Dong Yu},
      year={2025},
      eprint={2504.11456},
      archivePrefix={arXiv},
      primaryClass={cs.CL},
      url={https://arxiv.org/abs/2504.11456}, 
}

@misc{rein2023gpqagraduatelevelgoogleproofqa,
      title={GPQA: A Graduate-Level Google-Proof QA Benchmark}, 
      author={David Rein and Betty Li Hou and Asa Cooper Stickland and Jackson Petty and Richard Yuanzhe Pang and Julien Dirani and Julian Michael and Samuel R. Bowman},
      year={2023},
      eprint={2311.12022},
      archivePrefix={arXiv},
      primaryClass={cs.AI},
      url={https://arxiv.org/abs/2311.12022}, 
}

@misc{pteam2025supergpqascalingllmevaluation,
      title={SuperGPQA: Scaling LLM Evaluation across 285 Graduate Disciplines}, 
      author={P Team and Xinrun Du and Yifan Yao and Kaijing Ma and Bingli Wang and Tianyu Zheng and King Zhu and Minghao Liu and Yiming Liang and Xiaolong Jin and Zhenlin Wei and Chujie Zheng and Kaixin Deng and Shawn Gavin and Shian Jia and Sichao Jiang and Yiyan Liao and Rui Li and Qinrui Li and Sirun Li and Yizhi Li and Yunwen Li and David Ma and Yuansheng Ni and Haoran Que and Qiyao Wang and Zhoufutu Wen and Siwei Wu and Tyshawn Hsing and Ming Xu and Zhenzhu Yang and Zekun Moore Wang and Junting Zhou and Yuelin Bai and Xingyuan Bu and Chenglin Cai and Liang Chen and Yifan Chen and Chengtuo Cheng and Tianhao Cheng and Keyi Ding and Siming Huang and Yun Huang and Yaoru Li and Yizhe Li and Zhaoqun Li and Tianhao Liang and Chengdong Lin and Hongquan Lin and Yinghao Ma and Tianyang Pang and Zhongyuan Peng and Zifan Peng and Qige Qi and Shi Qiu and Xingwei Qu and Shanghaoran Quan and Yizhou Tan and Zili Wang and Chenqing Wang and Hao Wang and Yiya Wang and Yubo Wang and Jiajun Xu and Kexin Yang and Ruibin Yuan and Yuanhao Yue and Tianyang Zhan and Chun Zhang and Jinyang Zhang and Xiyue Zhang and Xingjian Zhang and Yue Zhang and Yongchi Zhao and Xiangyu Zheng and Chenghua Zhong and Yang Gao and Zhoujun Li and Dayiheng Liu and Qian Liu and Tianyu Liu and Shiwen Ni and Junran Peng and Yujia Qin and Wenbo Su and Guoyin Wang and Shi Wang and Jian Yang and Min Yang and Meng Cao and Xiang Yue and Zhaoxiang Zhang and Wangchunshu Zhou and Jiaheng Liu and Qunshu Lin and Wenhao Huang and Ge Zhang},
      year={2025},
      eprint={2502.14739},
      archivePrefix={arXiv},
      primaryClass={cs.CL},
      url={https://arxiv.org/abs/2502.14739}, 
}

@article{he2025nondeterminismThinky,
  author = {Horace He and Thinking Machines Lab},
  title = {Defeating Nondeterminism in LLM Inference},
  journal = {Thinking Machines Lab: Connectionism},
  year = {2025},
  note = {https://thinkingmachines.ai/blog/defeating-nondeterminism-in-llm-inference/},
  doi = {10.64434/tml.20250910}
}

@misc{yuan2025GiveMeFP32,
      title={Understanding and Mitigating Numerical Sources of Nondeterminism in LLM Inference}, 
      author={Jiayi Yuan and Hao Li and Xinheng Ding and Wenya Xie and Yu-Jhe Li and Wentian Zhao and Kun Wan and Jing Shi and Xia Hu and Zirui Liu},
      year={2025},
      eprint={2506.09501},
      archivePrefix={arXiv},
      primaryClass={cs.CL},
      url={https://arxiv.org/abs/2506.09501}, 
}

@misc{sglang_deterministic_2025,
  author       = {The SGLang Team},
  title        = {Towards Deterministic Inference in SGLang and Reproducible RL Training},
  howpublished = {\url{https://lmsys.org/blog/2025-09-22-sglang-deterministic/}},
  year         = {2025},
  month        = {September},
  day          = {22},
  note         = {LMSYS Org Blog}
}

@misc{fishman2025scalingfp8trainingtrilliontoken,
      title={Scaling FP8 training to trillion-token LLMs}, 
      author={Maxim Fishman and Brian Chmiel and Ron Banner and Daniel Soudry},
      year={2025},
      eprint={2409.12517},
      archivePrefix={arXiv},
      primaryClass={cs.LG},
      url={https://arxiv.org/abs/2409.12517}, 
}

@misc{peng2023fp8lm,
      title={FP8-LM: Training FP8 Large Language Models}, 
      author={Houwen Peng and Kan Wu and Yixuan Wei and Guoshuai Zhao and Yuxiang Yang and Ze Liu and Yifan Xiong and Ziyue Yang and Bolin Ni and Jingcheng Hu and Ruihang Li and Miaosen Zhang and Chen Li and Jia Ning and Ruizhe Wang and Zheng Zhang and Shuguang Liu and Joe Chau and Han Hu and Peng Cheng},
      year={2023},
      eprint={2310.18313},
      archivePrefix={arXiv},
      primaryClass={cs.LG},
      url={https://arxiv.org/abs/2310.18313}, 
}

@misc{kim2025inquiryfp8,
      title={An Inquiry into Datacenter TCO for LLM Inference with FP8}, 
      author={Jiwoo Kim and Joonhyung Lee and Gunho Park and Byeongwook Kim and Se Jung Kwon and Dongsoo Lee and Youngjoo Lee},
      year={2025},
      eprint={2502.01070},
      archivePrefix={arXiv},
      primaryClass={cs.LG},
      url={https://arxiv.org/abs/2502.01070}, 
}

@misc{SLIME_FP8Rollout,
  author       = {THUDM},
  title        = {Qwen3-4B Example — SLiME Documentation},
  year         = {2024},
  howpublished = {\url{https://github.com/THUDM/slime/blob/08118ecfa0570e838bf1299cf8ac9bacce8765ec/docs/en/examples/qwen3-4B.md\#bf16-training-with-fp8-inference/}},
  note         = {Accessed: 2025-10-30}
}

@misc{Nemo-RL_FP8Rollout,
  author       = {NVIDIA},
  title        = {FP8 Accuracy — NeMo-RL Documentation},
  year         = {2024},
  howpublished = {\url{https://github.com/NVIDIA-NeMo/RL/blob/bc24887c72a6e1b2699a228bc87c588546dfe6b7/docs/fp8.md\#accuracy/}},
  note         = {Accessed: 2025-10-30}
}

@misc{nvidia_transformer_engine,
  author       = {NVIDIA},
  title        = {Transformer Engine: A library for accelerating Transformer models on NVIDIA GPUs},
  howpublished = {\url{https://github.com/NVIDIA/TransformerEngine/}},
  year         = {2025},
  note         = {Version v2.8, accessed October 30, 2025}
}

@misc{ma2025reasoningmodelseffectivethinking,
      title={Reasoning Models Can Be Effective Without Thinking}, 
      author={Wenjie Ma and Jingxuan He and Charlie Snell and Tyler Griggs and Sewon Min and Matei Zaharia},
      year={2025},
      eprint={2504.09858},
      archivePrefix={arXiv},
      primaryClass={cs.AI},
      url={https://arxiv.org/abs/2504.09858}, 
}

@misc{huang2025qerlefficiencyquantizationenhanced,
      title={QeRL: Beyond Efficiency -- Quantization-enhanced Reinforcement Learning for LLMs}, 
      author={Wei Huang and Yi Ge and Shuai Yang and Yicheng Xiao and Huizi Mao and Yujun Lin and Hanrong Ye and Sifei Liu and Ka Chun Cheung and Hongxu Yin and Yao Lu and Xiaojuan Qi and Song Han and Yukang Chen},
      year={2025},
      eprint={2510.11696},
      archivePrefix={arXiv},
      primaryClass={cs.LG},
      url={https://arxiv.org/abs/2510.11696}, 
}




\end{document}